\documentclass{article}

\usepackage{microtype}
\usepackage{graphicx}
\usepackage{subcaption}
\usepackage{booktabs} 

\usepackage{hyperref}

\usepackage{amsmath}
\usepackage[T1]{fontenc}
\usepackage{graphicx}
\usepackage{booktabs}
\usepackage{amssymb}
\usepackage{adjustbox}
\usepackage{multirow}
\usepackage{subcaption}
\usepackage{float}
\usepackage{hyperref}
\usepackage{color}

\usepackage{amsmath}
\usepackage{esvect}

\usepackage{ifthen}  
\usepackage{xcolor}

\newcommand{\ShowComments}{yes}  

\ifthenelse{\equal{\ShowComments}{yes}}{
\newcommand{\Cen}[1]{{\color{blue}[Cen:#1]}}
\newcommand{\YC}[1]{{\color{brown}[YC: #1]}}
}{
\newcommand{\Cen}[1]{}
\newcommand{\YC}[1]{}
}

\usepackage{xcolor}

\newif\ifhighlightblue

\highlightbluefalse 

\newcommand{\bluetext}[1]{%
  {
    \ifhighlightblue\color{blue}\fi
    #1%
  }
}

\usepackage[accepted]{icml2026}

\usepackage{amsmath}
\usepackage{amssymb}
\usepackage{mathtools}
\usepackage{amsthm}

\usepackage[capitalize,noabbrev]{cleveref}

\theoremstyle{plain}

\theoremstyle{definition}

\theoremstyle{remark}

\usepackage[textsize=tiny]{todonotes}

\icmltitlerunning{Catastrophic Collapse in Vision-Language Models}

\begin{document}

\twocolumn[
  \icmltitle{Sparse Neuron Ablation Triggers Catastrophic Collapse \\ of the Language Core in Large Vision-Language Models}



\begin{icmlauthorlist}
  \icmlauthor{Cen Lu}{epfl,idiap}
  \icmlauthor{Yung-Chen Tang}{epfl,idiap}
  \icmlauthor{Andrea Cavallaro}{epfl}
\end{icmlauthorlist}

\icmlaffiliation{epfl}{École Polytechnique Fédérale de Lausanne (EPFL), Lausanne, Switzerland}
\icmlaffiliation{idiap}{Idiap Research Institute, Martigny, Switzerland}
\icmlcorrespondingauthor{Cen Lu}{cen.lu@epfl.ch}
  \icmlkeywords{Machine Learning, ICML}

  \vskip 0.3in
]



\printAffiliationsAndNotice{}  

\begin{abstract}
Large Vision-Language Models (LVLMs) have shown impressive multimodal understanding capabilities, yet the structures that sustain their functionality remain poorly understood from a mechanistic interpretability standpoint. We propose Consistently Activated Neurons (CAN), a progressive neuron ablation method to identify critical neurons whose removal triggers catastrophic collapse, and use it to investigate structural vulnerabilities in representative 7B LVLMs. Experiments reveal that catastrophic collapse can be triggered by ablating as few as four neurons in \texttt{LLaVA-1.5-7b-hf} and a few thousand in \texttt{InstructBLIP-vicuna-7b}, both representing a small fraction of model parameters. Notably, critical neurons are predominantly localized in the language model, particularly in its down-projection layer, rather than in the vision components. We also observe a consistent two-stage collapse pattern: initial expressive degradation followed by sudden, complete collapse. These findings reveal that LVLM functionality depends on a sparse subset of neurons concentrated in the language backbone, offering mechanistic insights into how their functionality is structured and where these models are most vulnerable.


\end{abstract}

\begin{figure}[t]

    \centering

    \includegraphics[width=1\linewidth]{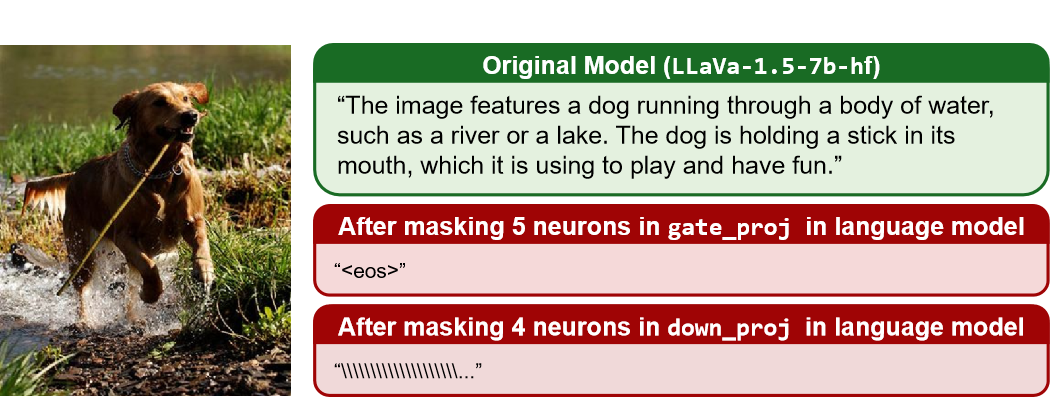}

    \caption{Given the prompt \texttt{Describe the object in this image}, the original \texttt{LLaVA-1.5-7b-hf} generates correct outputs (green box), but masking only 4–5 neurons in the language model causes catastrophic collapse (red boxes).}

    \label{fig:corrupted_output_example}

\end{figure}

\section{Introduction}

Large Vision-Language Models (LVLMs) (e.g.~LLaVA~\cite{DBLP:conf/nips/LiuLWL23a} and InstructBLIP~\cite{instructblip}) integrate visual perception with language to generate image descriptions or answer (visual) questions. Since these models are increasingly integrated into real-world applications \cite{ye2025multimodallargelanguagemodels,zhou2024visionlanguagemodelsautonomous}, assessing their robustness limits becomes important. 

Neuron-level analyses revealed that transformer models contain task-specific or language-specific neurons: certain neurons activate predominantly when processing syntax-related tasks, while others respond more strongly to specific languages~\cite{LLMLanguageNetwork,tang2024languagespecificneuronskeymultilingual}. Recent research on Large Language Models (LLMs) has also found  {\em super weights} that are decisive for model's functionality~\cite{qin2025achillesheelllmsaltering,yu2025superweightlargelanguage}. A tiny fraction of neurons can be crucial for safety mechanisms in LLMs: safety neurons have been identified that are responsible for model alignment and refusal behaviors, such as rejecting malicious prompts~\cite{chen2024finding,zhao2025understanding}. Inspired by neuroscience, researchers have developed interventional methods, such as neuron ablation~\cite{LLMLanguageNetwork}, that introduce {\em virtual lesions} to identify critical components~\cite{li2024optimal,meyes2019ablation}. Recent work has extended mechanistic understanding to multimodal settings, such as identifying domain-specific neurons specialized for medical images or document understanding~\cite{huo-etal-2024-mmneuron}. Manipulating these domain-specific neurons causes only modest performance changes on these domains~\cite{huo-etal-2024-mmneuron}.  However, existing works have largely focused on unimodal LLMs and overlooked the robustness limits of LVLMs from a mechanistic interpretability standpoint, for example the identification of subsets of neural units whose ablation triggers severe performance degradation or complete catastrophic collapse. This problem is important given the deployment of LVLMs in safety-critical applications such as autonomous driving \cite{zhou2024visionlanguagemodelsautonomous} and medical diagnosis \cite{LI2025102995}.

To characterize these structural vulnerabilities, we define \emph{catastrophic collapse} as a failure state in which the model's output distribution over its vocabulary becomes severely distorted, preventing the generation of any meaningful text across any valid input. Ablating a tiny subset of neurons (as few as four in \texttt{LLaVA-1.5-7b-hf}) in the language model can trigger catastrophic collapse, leading to degeneration such as repetitive or empty outputs. We refer to such neurons as \emph{critical neurons}.  While the number of neurons required varies across object categories and architectures, the underlying cause is consistent: removing only a tiny fraction of neurons suffices to collapse the tested models (Figure~\ref{fig:corrupted_output_example}).

In summary, our main contributions are:
\begin{itemize}
    \item  We propose the Consistently Activated Neurons (CAN) method \bluetext{as an empirical tool} for identifying \bluetext{candidate} critical neurons in 7B LVLMs across different images. Analyzing \texttt{LLaVA-1.5-7b-hf} and \texttt{InstructBLIP-vicuna-7b}, we generate component-specific rankings across vision encoder, cross-modal alignment module, and language model, and object-specific rankings, revealing neurons with the strongest activation patterns.

    \item  Leveraging CAN, we identify a very small subset of neurons that are disproportionately important for the model's output. For example, within our evaluation scope, masking only {four neurons} in \texttt{LLaVA-1.5-7b-hf}  suffices to collapse the model, showing structural vulnerability at the 7B parameter scale. 
    
    \item We localize critical neurons to the language model's feed-forward networks, showing that masking language-model neurons causes catastrophic collapse while vision-component ablations yield only degradation in visual understanding. This contrast reveals the language backbone as the primary structural bottleneck in LVLMs, providing a mechanistic interpretability perspective on multimodal vulnerability.

\end{itemize}

\section{Related work}

{\em Neuron ablation} is a powerful technique for understanding the internal mechanisms and functional specialization within neural networks~\cite{LLMLanguageNetwork}. For example, the {\em neural erosion} method~\cite{NeuralErosion} applies synaptic pruning and neuron deactivation in LLMs, showing that LLMs lose cognitive abilities in a hierarchical manner: mathematical reasoning first, followed by linguistic capabilities. By replacing neuron activations with their mean values across samples, researchers have identified specialized entropy neurons and token frequency neurons that impact model confidence rather than direct predictions, revealing calibration circuits across architectures~\cite{confidenceneuron}. Additionally, recent work applies neuron ablation to evaluate role-playing prompts in medical LLMs, showing that different role assignments activate similar groups of neurons and affect only linguistic features rather than core reasoning~\cite{medicalneuron}. These studies focus on identifying functional neurons and overlook the ablation thresholds that may lead to catastrophic collapse.

While traditional neuron ablation identifies neurons critical for general model performance, {\em safety neuron identification} targets neurons responsible for safety alignment, ensuring models reject malicious prompts and generate safe responses. Contrastive methods to identify safety neurons in LLMs showed that these neurons are both sparse and effective, with only 5\% of neurons responsible for 90\% of safety performance~\cite{chen2024finding}. Safety neurons constitute fewer than 1\% of all parameters and reside in self-attention layers~\cite{zhao2025understanding}. Complementing neuron-level analyses, Zhou et al.~\cite{zhou2025role} showed that ablating safety-related attention heads degrades safety capabilities. While these safety-focused methods rely on contrastive activation patterns between aligned and unaligned models, we develop a neuroscience-inspired approach to identify critical neurons in LVLMs through activation-gradient ranking to investigate failure risks, as illustrated in Section~\ref{sec:can}.

\section{Problem formulation}

Let $\mathcal{M}$ be an LVLM that processes a visual input $\mathbf{v}$ and a textual prompt $\mathbf{t}$ to generate  $\mathcal{M}(\mathbf{v}, \mathbf{t})$, the output response. Let $\mathcal{N} = \{n_{\ell,j} \mid \ell \in \mathcal{L}, j \in \mathcal{J}_\ell\}$ denote the set of neurons across all layers $\mathcal{L}$ and indices $\mathcal{J}_\ell$, where $\ell$ is the layer index and $j$ is the position index. Given a dataset $\mathcal{D} = \{(\mathbf{v}_i, \mathbf{t})\}_{i=1}^M$ containing $M$ vision-language pairs with a fixed textual prompt, our objective is to identify a critical neuron subset $\mathcal{C}^* \subset \mathcal{N}$, with $|\mathcal{C}^*| \ll |\mathcal{N}|$ that satisfies:
\vspace{-2em}

\begin{equation}
    \mathcal{C}^* = \arg\min_{\mathcal{C} \subset \mathcal{N}} |\mathcal{C}| \quad \text{s.t.} \quad \frac{1}{M}\sum_{i=1}^M \Delta_{\text{perf}}(\mathcal{M} | \mathcal{C}, \mathbf{v}_i, \mathbf{t}) \geq \tau,
\end{equation}
where $\Delta_{\text{perf}}(\mathcal{M} | \mathcal{C}, \mathbf{v}_i, \mathbf{t})$ quantifies performance degradation of model $\mathcal{M}$ when neurons in $\mathcal{C}$ are ablated on input $(\mathbf{v}_i, \mathbf{t})$, and $\tau$ is a threshold indicating catastrophic failure. We define  $\tau$ through joint criteria based on perplexity~\cite{ppl} and CLIP score~\cite{clipscore}, as described in Section~\ref{sec:clip and ppl}. 

\begin{figure*}[t]
    \centering
    \includegraphics[width=0.8\linewidth]{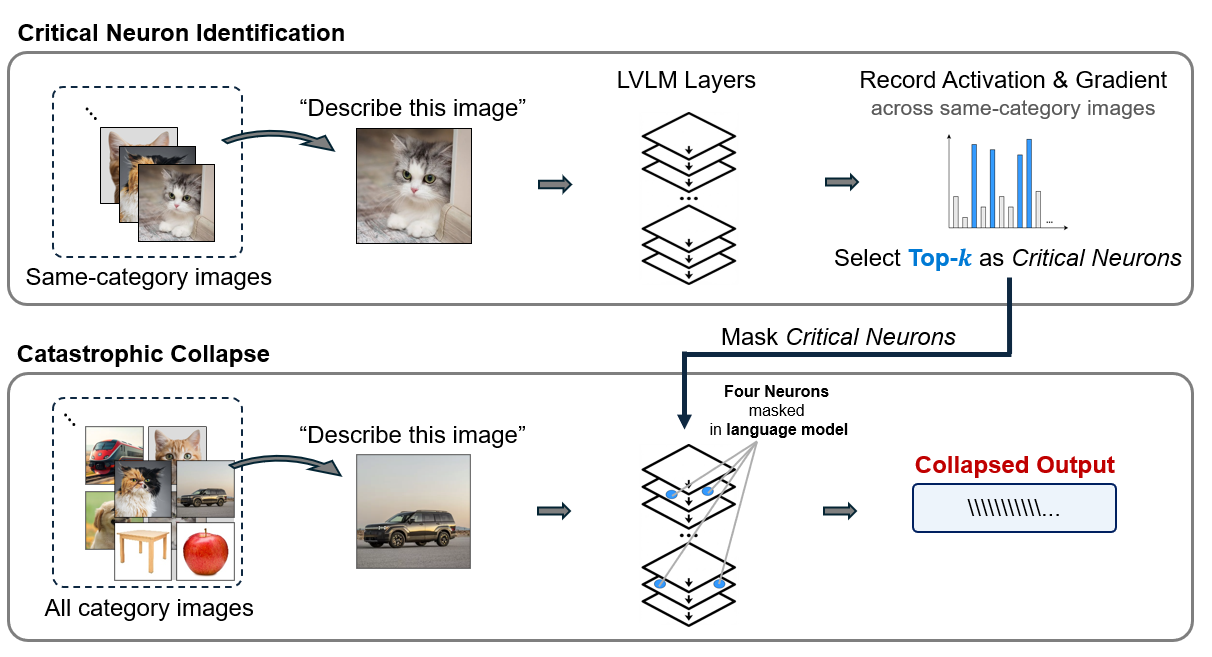}
    \caption{CAN first ranks neurons by importance combining activation and gradient magnitudes. Next, masking the top-$k$ critical neurons causes catastrophic collapse, with the generation of meaningless outputs regardless of the input.}
    \label{fig:pipeline}
\end{figure*}

\section{CAN: Consistently Activated Neurons}

\label{sec:can}
To identify critical neurons in LVLMs, we draw inspiration from lesion-function analyses in neuroscience~\cite{bates2003voxel}. \emph{Lesion-symptom mapping} establishes functional specificity by associating selective cognitive impairments with localized brain damage~\cite{kimberg2007power,meyer2016voxel}. Translating this principle to neural networks, we hypothesize that LVLMs contain sparse subsets of neurons that are disproportionately critical for a broad range of tasks~\cite{qin2025achillesheelllmsaltering,yu2025superweightlargelanguage}. Specifically, we employ neuron ablation techniques to validate the importance of identified neurons, aligning with neuroscience practices such as \emph{virtual lesioning} (e.g., Transcranial Magnetic Stimulation \cite{hallett2007transcranial}). Furthermore, we use random neuron ablation as a control to mimic the practice of comparing targeted lesions to random ones in neuroscience experiments~\cite{rorden2004using}.

Inspired by activation-based saliency~\cite{simonyan2014deepinsideconvolutionalnetworks} and gradient-based pruning criteria~\cite{lee2019snipsingleshotnetworkpruning,lin2018focallossdenseobject}, we propose CAN. Unlike these methods, which target performance-preserving pruning in unimodal settings, CAN is designed to identify collapse-triggering neurons across the multimodal components of LVLMs. CAN quantifies neuron importance through \emph{activation magnitude} and \emph{gradient sensitivity}. The activation-based importance contribution focuses on how strongly neurons respond to inputs (forward pass). The gradient-based importance contribution focuses on how changes to neurons would affect the loss, revealing their impact on model predictions (backward pass). The method is explained in Figure~\ref{fig:pipeline}.

For our localization protocol, we extract activation profiles across all layers during forward propagation. For each neuron at layer $\ell$ and index $j$, we record its activation values across all samples in $\mathcal{D}_{\mathrm{cal}}$, the calibration set used for neuron importance scoring:
\begin{equation}
    \mathbf{A}_{\ell,j} = \{a_{\ell,j}(\mathbf{v}_i, \mathbf{t}) \mid (\mathbf{v}_i, \mathbf{t}) \in \mathcal{D}_{\mathrm{cal}}\},
\end{equation}
where $|\mathbf{A}_{\ell,j}| = M$ and $a_{\ell,j} \in \mathbb{R}$.

To quantify neuron importance, we adopt a scoring mechanism combining activation magnitude with gradient sensitivity, related to criteria in pruning literature~\cite{molchanov2017pruning}. For each neuron at layer $\ell$ and index $j$, we compute:
\begin{equation}
    \mathcal{I}_{\ell,j} = \frac{1}{M}\sum_{(\mathbf{v}_i, \mathbf{t}) \in \mathcal{D}_{\mathrm{cal}}} \left|\frac{\partial \mathcal{L}(\mathbf{v}_i, \mathbf{t})}{\partial a_{\ell,j}(\mathbf{v}_i, \mathbf{t})}\right|^\alpha \cdot |a_{\ell,j}(\mathbf{v}_i, \mathbf{t})|,
    \label{eq:importance_score}
\end{equation}
where $\frac{\partial \mathcal{L}}{\partial a_{\ell,j}}$ represents the gradient sensitivity of the loss with respect to neuron activation, and the hyperparameter $\alpha \in [0,1]$ controls the balance between the gradient term and the activation magnitude. By selectively combining the magnitude and gradient signals through $\alpha$, we obtain a flexible measure of neuron criticality across LVLMs.

Next, we rank all neurons by their importance scores in descending order. The top-$k$ neurons are then selected as critical neurons:
\begin{equation}
    \mathcal{C}_k = \{n_{\ell,j} : \mathcal{I}_{\ell,j} \text{ ranks in top-}k \text{ globally}\},
\end{equation}
where the hyperparameter $k$ is determined through iterative masking by increasing $k$ until a catastrophic performance degradation occurs. This identifies the number of critical neurons $k^*$ that suffices to trigger catastrophic collapse for each specific model and component.

\section{Assessing performance degradation }
\label{sec:clip and ppl}

We evaluate performance degradation using perplexity~\cite{ppl}, to assess the language modeling capability, and CLIP Score~\cite{clipscore}, to verify image-text alignment. Perplexity measures a model's ability to maintain probability distribution over the original model's output, evaluating whether the masked model can still follow the base model's linguistic patterns. The CLIP score measures semantic alignment between generated text and visual content. A low CLIP score with stable perplexity can arise from two distinct failure modes, namely \emph{perceptual failure} and \emph{expressive degradation}.  With a \emph{perceptual failure} the model generates fluent but irrelevant descriptions (e.g., describing a cat when shown a car), indicating damage of vision neurons and loss of visual grounding, while preserving linguistic capabilities. With an \emph{expressive degradation} the model outputs malformed text, suggesting damage to the language generation component.

We measure the {\em perplexity degradation} on a validation set $\mathcal{D}_{\text{val}} = \{(\mathbf{v}_i, \mathbf{t})\}_{i=1}^{M_{\text{val}}}$, which consists of validation image-prompt pairs. For each input pair $(\mathbf{v}_i, \mathbf{t}) \in \mathcal{D}_{\text{val}}$ and masking configuration $\mathcal{C}_k$, the model generates a token sequence $\mathbf{y}^{(i)} = (y^{(i)}_1, \ldots, y^{(i)}_{L^{(i)}})$ of length $L^{(i)}$. The perplexity is:
\begin{equation}
    \text{PPL}(\mathbf{y}^{(i)} | \mathbf{v}_i, \mathbf{t}) = \exp\!\left(-\tfrac{1}{L^{(i)}}\sum_{j=1}^{L^{(i)}} \log p(y^{(i)}_j | y^{(i)}_{1:j-1}, \mathbf{v}_i, \mathbf{t})\right)\!.
\end{equation}

We measure the incremental log-ratio of perplexities to quantify the sudden deviation when masking additional neurons:
\begin{equation}
    \Delta_{\text{PPL}}(\mathcal{C}_k) = \frac{1}{M_{\text{val}}}\sum_{i=1}^{M_{\text{val}}} \log_{10}\left(\frac{\text{PPL}(\mathcal{M} | \mathcal{C}_k, \mathbf{v}_i, \mathbf{t})}{\text{PPL}(\mathcal{M} | \mathcal{C}_{k-\Delta k}, \mathbf{v}_i, \mathbf{t})}\right),
\end{equation}
where $\text{PPL}(\mathcal{M} | \mathcal{C}_k, \mathbf{v}_i, \mathbf{t})$ is the perplexity of model $\mathcal{M}$ with neurons in $\mathcal{C}_k$ masked, $\text{PPL}(\mathcal{M} | \mathcal{C}_{k-\Delta k}, \mathbf{v}_i, \mathbf{t})$ is the perplexity at the previous masking step, and $\Delta k$ is the masking step size. We use log-scale to capture order-of-magnitude changes in perplexity. This metric evaluates the incremental degradation in language modeling capability: larger values indicate a sudden collapse when masking the current set of neurons compared to the previous step.

The CLIP score~\cite{clipscore} measures semantic coherence between visual inputs and generated descriptions. We compute the CLIP score on the validation set:

\begin{equation}
    \text{CLIP}(\mathcal{C}_k) = \frac{1}{M_{\text{val}}}\sum_{i=1}^{M_{\text{val}}} \text{CLIP}(\mathbf{v}_i, (\mathcal{M} | \mathcal{C}_k, \mathbf{v}_i, \mathbf{t})),
\end{equation}
where $\text{CLIP}(\mathbf{v}_i, (\mathcal{M} | \mathcal{C}_k, \mathbf{v}_i, \mathbf{t}))$ is the cosine similarity between the CLIP embedding of image $\mathbf{v}_i$ and the text generated by model $\mathcal{M}$ with neurons in $\mathcal{C}_k$ masked on input $(\mathbf{v}_i, \mathbf{t})$. The lower the CLIP score, the worse the semantic alignment between the visual input and the generated text.

We determine the critical neuron set size $k$ in a data-driven manner by greedy search  with step size $\Delta k$. We increase $k$ until catastrophic failure occurs:

\begin{equation}
    k^* = \min\{k : \Delta_{\text{PPL}}(\mathcal{C}_k) \geq \tau_{\text{PPL}} \text{ and } \text{CLIP}(\mathcal{C}_k) \leq \tau_{\text{CLIP}}\},
\end{equation}
where $\tau_{\text{PPL}}$ and $\tau_{\text{CLIP}}$ are failure thresholds. A \emph{complete collapse} occurs when $\Delta_{\text{PPL}} \geq \tau_{\text{PPL}}$ and $\text{CLIP} \leq \tau_{\text{CLIP}}$, illustrating catastrophic collapse of core model capabilities. We set $\tau_{\text{PPL}} = 1$ (indicating one order of magnitude degradation) and $\tau_{\text{CLIP}} = 22$ based on empirical failure patterns (see details in Section~\ref{sec:critical neuron location}).

\section{Experiments}
\noindent\textbf{Architecture-Specific Neuron Localization.} We adapt the neuron identification to the specific components of LLaVA~\cite{llava} and InstructBLIP~\cite{instructblip}, as CAN is applicable across different LVLM architectures. For {LLaVA}, we identify neurons in: vision encoder, 
multimodal projector, and language model. For {InstructBLIP}: vision encoder, Q-Former with dual attention over learnable queries, and language model. In Eq.~\eqref{eq:importance_score}, we empirically set $\alpha = 1$ for InstructBLIP since gradient information is essential due to the Q-Former's complex attention routing, and $\alpha = 0$ for LLaVA,  as its simpler projector architecture makes activation magnitude alone a stable and sufficient signal for critical neuron identification.

\noindent\textbf{Experimental setup.} We conduct experiments with \texttt{InstructBLIP-vicuna-7b} and \texttt{LLaVA-1.5-7b-hf} on a dataset $\mathcal{D}_{\mathrm{cal}}$ of 300 images for each of 10 object categories from the \texttt{ILSVRC2013} dataset~\cite{deng2009imagenet}. All experiments were executed on an H100 GPU using fp32 precision. To generate ranked neuron lists, we use images from the 10  categories with \texttt{Describe the object in this image} as prompt $\mathbf{t}$. We found this simple prompt to be effective for critical neuron identification after testing different prompts, including more detailed instructions. We record the activation magnitude of each neuron across all model components after generating the first token, then rank neurons using CAN across all image categories. For each object category, we evaluate the progressively masked model using the additional 100 test images from the same category as $\mathcal{D}_{\mathrm{val}}$ .  
In the following, we address the minimum number of critical neurons, their stability across inputs, the effect of masking position, and their localization.


\begin{table*}[t]
\raggedright
\caption{Comparison of catastrophic collapse thresholds between \texttt{LLaVA-1.5-7b-hf} and \texttt{InstructBLIP-vicuna-7b} across different object categories by ablating neurons in \texttt{gate\_proj}. Each row shows results averaged over 100 test images. Progressive steps of 100 are used for the latter to find an approximate collapse threshold in \texttt{InstructBLIP-vicuna-7b}.}
\label{tab:collapse_comparison}

{\fontsize{14pt}{18pt}\selectfont
\begin{adjustbox}{width=1\textwidth}
\begin{tabular}{ccccccc}
\toprule
\multirow{2}{*}{Object} & \multicolumn{3}{c}{\textbf{LLaVA-1.5-7b-hf}} & \multicolumn{3}{c}{\textbf{InstructBLIP-vicuna-7b}} \\
\cmidrule(lr){2-4} \cmidrule(lr){5-7}
 & Neurons & PPL Change ($\downarrow$) & CLIP Score Change ($\uparrow$) &
 Neurons & PPL Change ($\downarrow$) & CLIP Score Change ($\uparrow$) \\
 & (\#) & (Orig. $\rightarrow$ Masked, \textbf{Change}$^{\dagger}$) &
 (Orig. $\rightarrow$ Masked, \textbf{\% Change}$^{\dagger}$)
 & (\#) & (Orig. $\rightarrow$ Masked, \textbf{Change}$^{\dagger}$) &
 (Orig. $\rightarrow$ Masked, \textbf{\% Change}$^{\dagger}$) \\
\midrule
dog & 5 & 2.20 $\rightarrow$ 8.82$\times10^{3}$ (\textbf{$\times$4009}) &
30.35 $\rightarrow$ NaN$^{\ddagger}$ (\textbf{-100\%}) &
3100 & 1.75 $\rightarrow$ 1.38$\times10^{2}$ (\textbf{$\times$79}) &
31.44 $\rightarrow$ 20.02 (\textbf{-36\%}) \\
cat & 5 & 2.54 $\rightarrow$ 7.34$\times10^{3}$ (\textbf{$\times$2890}) &
31.44 $\rightarrow$ NaN$^{\ddagger}$ (\textbf{-100\%}) &
2700 & 1.68 $\rightarrow$ 1.15$\times10^{2}$ (\textbf{$\times$69}) &
29.95 $\rightarrow$ 21.05 (\textbf{-30\%}) \\
car & 5 & 2.26 $\rightarrow$ 8.56$\times10^{3}$ (\textbf{$\times$3788}) &
30.80 $\rightarrow$ NaN$^{\ddagger}$ (\textbf{-100\%}) &
1200 & 2.02 $\rightarrow$ 7.59$\times10^{1}$ (\textbf{$\times$38}) &
29.58 $\rightarrow$ 19.04 (\textbf{-36\%}) \\
pencil box & 5 & 3.25 $\rightarrow$ 7.21$\times10^{3}$ (\textbf{$\times$2218}) &
29.52 $\rightarrow$ NaN$^{\ddagger}$ (\textbf{-100\%}) &
2000 & 1.74 $\rightarrow$ 9.06$\times10^{1}$ (\textbf{$\times$52}) &
29.68 $\rightarrow$ 21.25 (\textbf{-28\%}) \\
sofa & 5 & 2.13 $\rightarrow$ 9.28$\times10^{3}$ (\textbf{$\times$4357}) &
31.18 $\rightarrow$ NaN$^{\ddagger}$ (\textbf{-100\%}) &
2400 & 1.75 $\rightarrow$ 8.60$\times10^{1}$ (\textbf{$\times$49}) &
32.67 $\rightarrow$ 20.28 (\textbf{-38\%}) \\
table & 5 & 2.08 $\rightarrow$ 1.02$\times10^{4}$ (\textbf{$\times$4904}) &
29.72 $\rightarrow$ NaN$^{\ddagger}$ (\textbf{-100\%}) &
1300 & 2.06 $\rightarrow$ 6.31$\times10^{1}$ (\textbf{$\times$31}) &
30.99 $\rightarrow$ 19.41 (\textbf{-37\%}) \\
train & 5 & 2.80 $\rightarrow$ 9.83$\times10^{3}$ (\textbf{$\times$3511}) &
28.16 $\rightarrow$ NaN$^{\ddagger}$ (\textbf{-100\%}) &
10600 & 2.34 $\rightarrow$ 3.55$\times10^{2}$ (\textbf{$\times$152}) &
29.74 $\rightarrow$ 18.94 (\textbf{-36\%}) \\
apple & 5 & 2.08 $\rightarrow$ 9.76$\times10^{3}$ (\textbf{$\times$4692}) &
31.64 $\rightarrow$ NaN$^{\ddagger}$ (\textbf{-100\%}) &
2300 & 1.66 $\rightarrow$ 7.07$\times10^{1}$ (\textbf{$\times$43}) &
32.30 $\rightarrow$ 21.02 (\textbf{-35\%}) \\
orange & 5 & 2.38 $\rightarrow$ 9.99$\times10^{3}$ (\textbf{$\times$4199}) &
32.06 $\rightarrow$ NaN$^{\ddagger}$ (\textbf{-100\%}) &
1900 & 1.65 $\rightarrow$ 7.41$\times10^{1}$ (\textbf{$\times$45}) &
31.49 $\rightarrow$ 20.96 (\textbf{-33\%}) \\
cup & 5 & 2.99 $\rightarrow$ 7.62$\times10^{3}$ (\textbf{$\times$2549}) &
30.46 $\rightarrow$ NaN$^{\ddagger}$ (\textbf{-100\%}) &
1400 & 1.64 $\rightarrow$ 1.20$\times10^{2}$ (\textbf{$\times$73}) &
32.21 $\rightarrow$ 20.04 (\textbf{-38\%}) \\
\bottomrule
\multicolumn{7}{l}{ $^{\dagger}$Orig. and Masked denote metrics from the original model and after masking critical neurons at collapse.} \\
\multicolumn{7}{l}{ $^{\ddagger}$NaN occurs when the model outputs only end-of-sentence tokens, making CLIP score computation infeasible. For \% Change calculation, NaN is}\\
\multicolumn{7}{l}{\ treated as 0 (i.e., -100\%).}
\end{tabular}
\end{adjustbox}
}

\end{table*}

\begin{figure*}[t]
    \centering
    \includegraphics[width=0.8\linewidth]{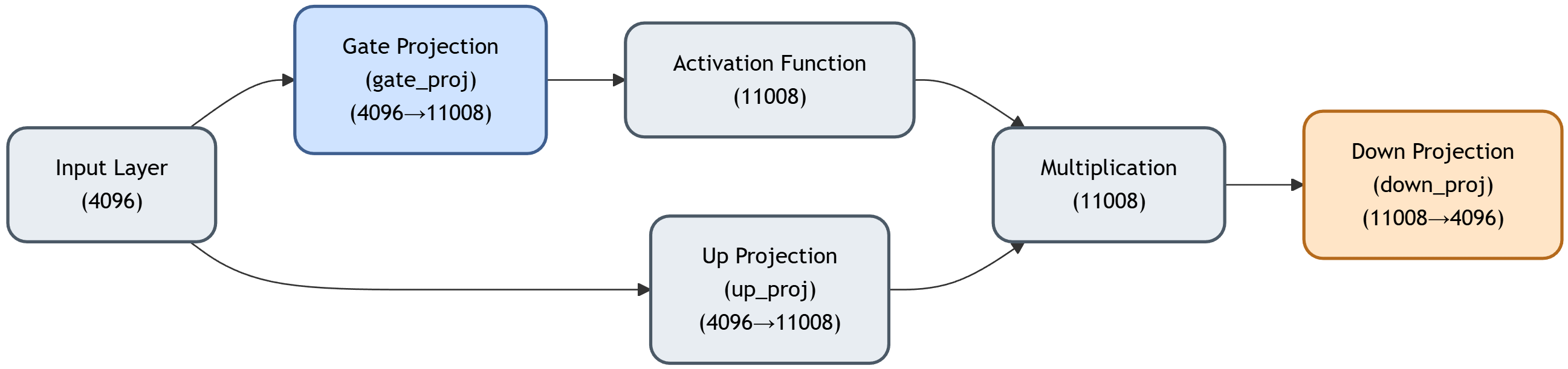}

    \caption{Architecture of the Feed-Forward Network (FFN) module in the transformer layer for \texttt{LLaVA-1.5-7b-hf} and \texttt{InstructBLIP-vicuna-7b}. Numbers in parentheses indicate tensor dimensions. The identification and masking of critical neurons in the blue component (\texttt{gate\_proj}) and the orange component (\texttt{down\_proj}) are discussed in Section~\ref{sec:Q1} and Section~\ref{sec:q3}, respectively.}
    \label{fig:mlp}
\end{figure*}

\begin{figure}[t]
    \centering
    \begin{subfigure}[t]{0.8\linewidth}
        \centering
        \includegraphics[width=\linewidth]{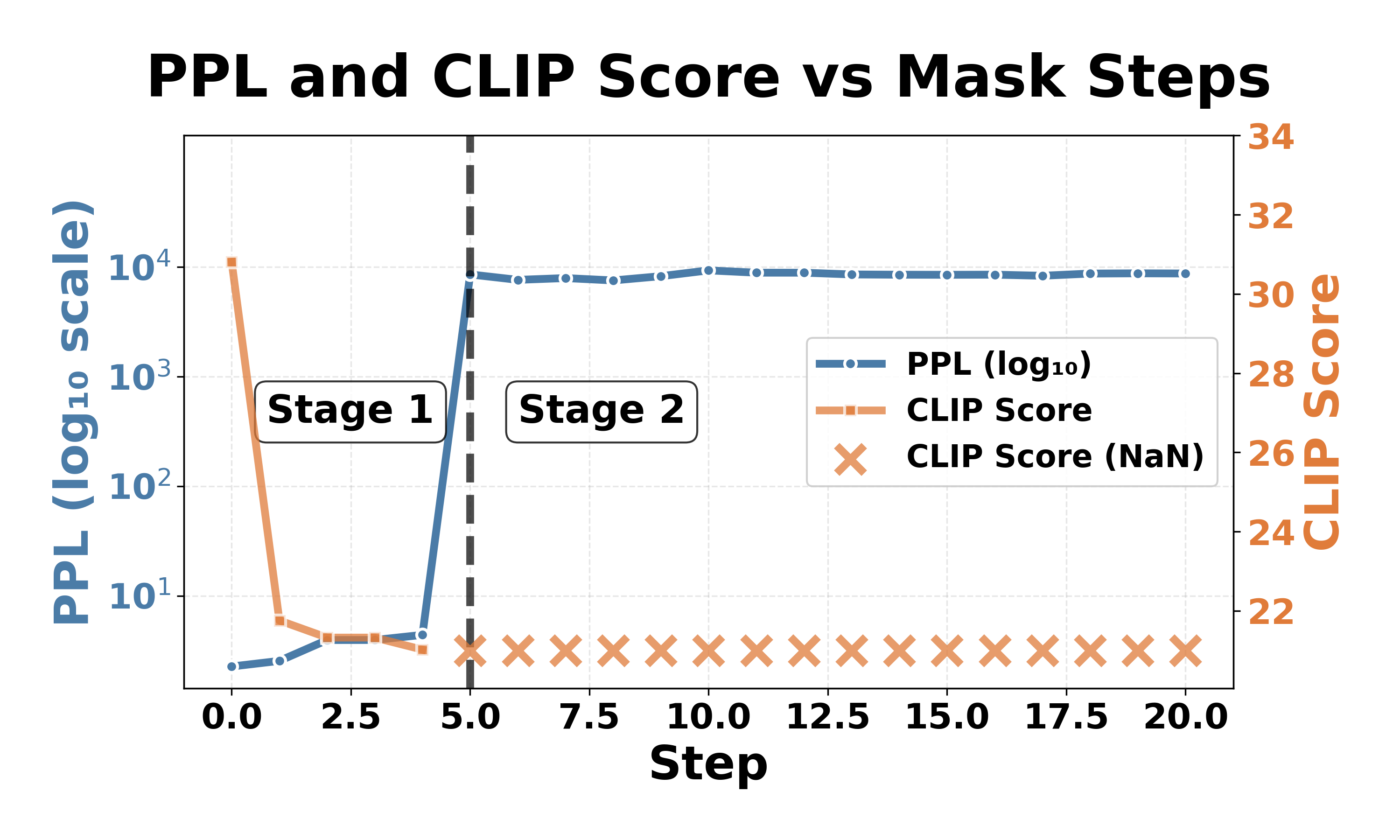}
        \caption{\texttt{LLaVA-1.5-7b-hf}}
        \label{fig:llava_case_study}
    \end{subfigure}
    
    \vspace{0.5em}
    
    \begin{subfigure}[t]{0.8\linewidth}
        \centering
        \includegraphics[width=\linewidth]{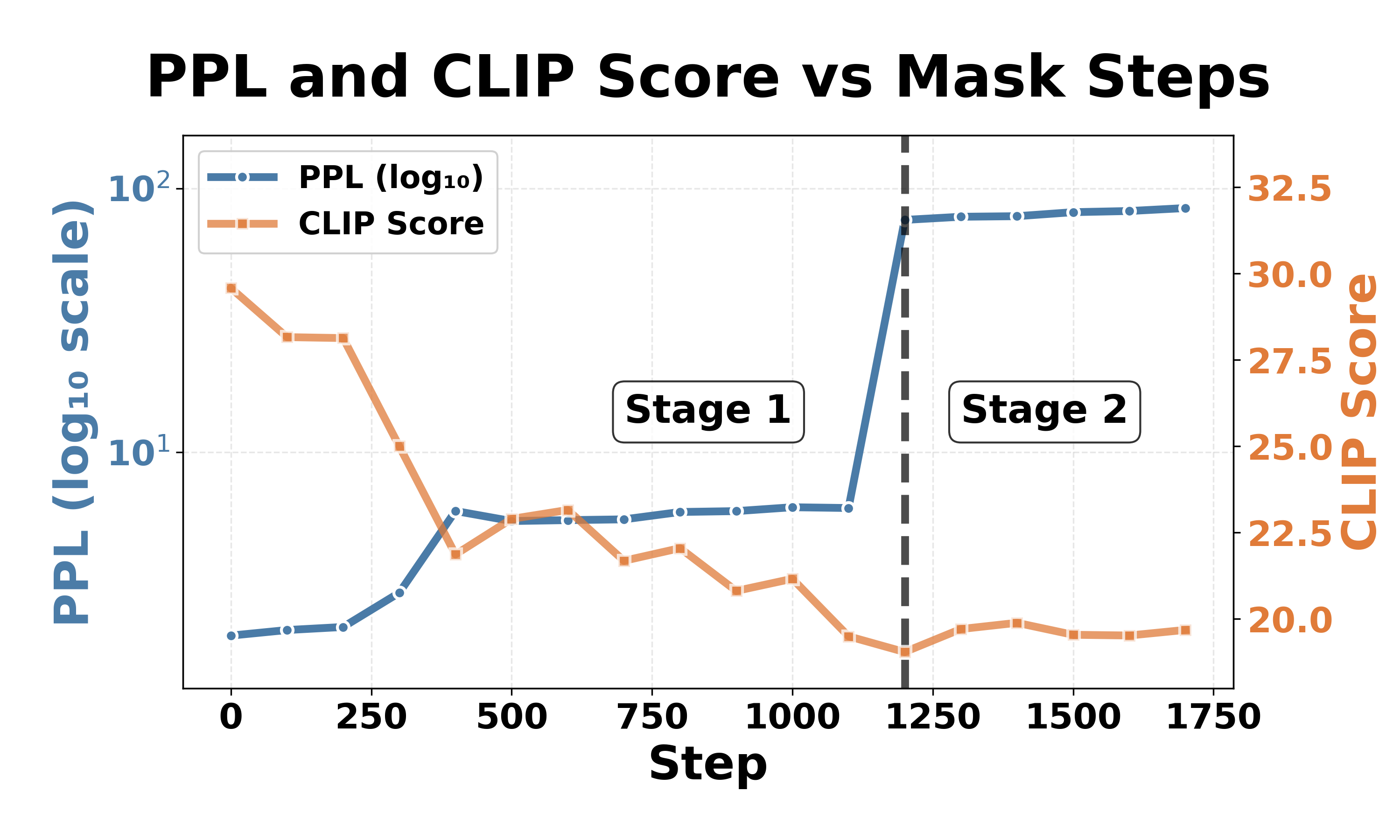}
        \caption{\texttt{InstructBLIP-vicuna-7b}}
        \label{fig:instructblip_case_study}
    \end{subfigure}
    \caption{Comparison of PPL and CLIP score changes during progressive masking of neurons in \texttt{gate\_proj} from FFN using car images. Stage 1 indicates expressive degradation and Stage 2 indicates complete collapse, as explained in Section~\ref{sec:Q1}.}
    \label{fig:comparison_case_study}
\end{figure}

\subsection{Number of neurons for catastrophic collapse}
\label{sec:Q1}
We progressively masked critical parts in \texttt{gate\_proj} (see Figure~\ref{fig:mlp}) from the feed-forward network (FFN) module in the language model of LVLMs. Our results on \texttt{LLaVA-1.5-7b-hf} show a consistent pattern of catastrophic collapse across different object categories. As shown in Table~\ref{tab:collapse_comparison}, \texttt{LLaVA-1.5-7b-hf} requires only five neurons to trigger catastrophic failure across all 10 object categories, with perplexity increases exceeding two orders of magnitude and CLIP scores becoming undefined (NaN) as the model outputs only end-of-sentence tokens, making CLIP score computation infeasible. These five neurons remain consistent regardless of the object category, with three neurons located in layer 1 and two neurons in layer 30 (out of 32 layers, zero-indexed), suggesting that neurons at the beginning and end of \texttt{LLaVA-1.5-7b-hf} may play critical roles in maintaining model stability. In contrast, \texttt{InstructBLIP-vicuna-7b} is more  robust as it requires at least 1200 neurons (still representing an extremely small fraction of $0.17\times10^{-6}$ of the model parameters), depending on the object category. Notably, once catastrophic collapse is triggered, it occurs regardless of whether the input is pure text or image–text pairs.

\begin{table*}[h]
\raggedright
\caption{Collapse induced by masking neurons in the \texttt{down\_proj} component of \texttt{LLaVA-1.5-7b-hf} and \texttt{InstructBLIP-vicuna-7b}. Progressive steps of 100 are used for the latter to find an approximate collapse threshold. The collapse manifests as perplexity increases of at least three orders of magnitude and a significant drop in CLIP Score to below 22.}
\label{tab:collapse_downproj}

{\fontsize{14pt}{18pt}\selectfont
\begin{adjustbox}{width=1\textwidth}
\begin{tabular}{ccccccc}
\toprule
\multirow{2}{*}{Object} & \multicolumn{3}{c}{\textbf{LLaVA-1.5-7b-hf}} & \multicolumn{3}{c}{\textbf{InstructBLIP-vicuna-7b}} \\
\cmidrule(lr){2-4} \cmidrule(lr){5-7}
 & Neurons & PPL Change ($\downarrow$) & CLIP Score Change ($\uparrow$) & Neurons & PPL Change ($\downarrow$) & CLIP Score Change ($\uparrow$) \\
 & (\#) & (Orig. $\rightarrow$ Masked, \textbf{Change}$^{\dagger}$) & (Orig. $\rightarrow$ Masked, \textbf{\% Change}$^{\dagger}$) & (\#) & (Orig. $\rightarrow$ Masked, \textbf{Change}$^{\dagger}$) & (Orig. $\rightarrow$ Masked, \textbf{\% Change}$^{\dagger}$) \\
\midrule
dog & 4 & 2.20 $\rightarrow$ 1.44$\times10^{4}$ (\textbf{$\times$6545}) & 30.35 $\rightarrow$ 20.26 (\textbf{-33\%}) & 2400 & 1.75 $\rightarrow$ 3.78$\times10^{3}$ (\textbf{$\times$2160}) & 31.44 $\rightarrow$ 20.09 (\textbf{-36\%}) \\
cat & 4 & 2.54 $\rightarrow$ 1.21$\times10^{4}$ (\textbf{$\times$4764}) & 31.44 $\rightarrow$ 20.53 (\textbf{-35\%}) & 2500 & 1.68 $\rightarrow$ 2.93$\times10^{3}$ (\textbf{$\times$1744}) & 29.95 $\rightarrow$ 21.61 (\textbf{-28\%}) \\
car & 4 & 2.26 $\rightarrow$ 1.53$\times10^{4}$ (\textbf{$\times$6770}) & 30.80 $\rightarrow$ 20.41 (\textbf{-34\%}) & 1000 & 2.02 $\rightarrow$ 4.65$\times10^{3}$ (\textbf{$\times$2302}) & 29.58 $\rightarrow$ 18.98 (\textbf{-36\%}) \\
pencil box & 4 & 3.25 $\rightarrow$ 1.61$\times10^{4}$ (\textbf{$\times$4954}) & 29.52 $\rightarrow$ 19.48 (\textbf{-34\%}) & 1300 & 1.74 $\rightarrow$ 9.01$\times10^{2}$ (\textbf{$\times$518}) & 29.68 $\rightarrow$ 19.81 (\textbf{-33\%}) \\
sofa & 4 & 2.13 $\rightarrow$ 1.23$\times10^{4}$ (\textbf{$\times$5775}) & 31.18 $\rightarrow$ 19.58 (\textbf{-37\%}) & 2000 & 1.75 $\rightarrow$ 2.34$\times10^{3}$ (\textbf{$\times$1337}) & 32.67 $\rightarrow$ 19.51 (\textbf{-40\%}) \\
table & 4 & 2.08 $\rightarrow$ 1.54$\times10^{4}$ (\textbf{$\times$7404}) & 29.72 $\rightarrow$ 19.25 (\textbf{-35\%}) & 1000 & 2.06 $\rightarrow$ 2.34$\times10^{3}$ (\textbf{$\times$1136}) & 30.99 $\rightarrow$ 19.42 (\textbf{-37\%}) \\
train & 4 & 2.80 $\rightarrow$ 2.21$\times10^{4}$ (\textbf{$\times$7893}) & 28.16 $\rightarrow$ 20.48 (\textbf{-27\%}) & 14400 & 2.34 $\rightarrow$ 1.08$\times10^{4}$ (\textbf{$\times$4615}) & 29.74 $\rightarrow$ 18.47 (\textbf{-38\%}) \\
apple & 4 & 2.08 $\rightarrow$ 1.59$\times10^{4}$ (\textbf{$\times$7644}) & 31.64 $\rightarrow$ 20.50 (\textbf{-35\%}) & 2100 & 1.66 $\rightarrow$ 2.84$\times10^{3}$ (\textbf{$\times$1711}) & 32.30 $\rightarrow$ 20.70 (\textbf{-36\%}) \\
orange & 4 & 2.38 $\rightarrow$ 1.94$\times10^{4}$ (\textbf{$\times$8151}) & 32.06 $\rightarrow$ 21.17 (\textbf{-34\%}) & 2800 & 1.65 $\rightarrow$ 2.64$\times10^{3}$ (\textbf{$\times$1604}) & 31.49 $\rightarrow$ 21.70 (\textbf{-30\%}) \\
cup & 4 & 2.99 $\rightarrow$ 1.08$\times10^{4}$ (\textbf{$\times$3612}) & 30.46 $\rightarrow$ 20.52 (\textbf{-33\%}) & 1100 & 1.64 $\rightarrow$ 1.87$\times10^{3}$ (\textbf{$\times$1140}) & 32.21 $\rightarrow$ 21.15 (\textbf{-34\%}) \\
\bottomrule
\multicolumn{7}{l}{$^{\dagger}$Orig. and Masked denote metrics before and after collapse induced by masking critical neurons.}
\end{tabular}
\end{adjustbox}
}

\end{table*}

\subsection{Stability}

Figure~\ref{fig:comparison_case_study} illustrates the collapse progression for both models on the car category. Despite different thresholds, both models exhibit similar two-stage patterns, with an expressive degradation followed by a complete collapse. 

\noindent{\bf Stage 1:  expressive degradation.} For \texttt{LLaVA-1.5-7b-\\hf} (Figure~\ref{fig:llava_case_study}, Steps 0--4), CLIP scores gradually decrease from 31 to 21 while perplexity remains stable, indicating declining output quality with preserved comprehension, similar to \textit{expressive aphasia} \cite{simmons2005outcome} in human cognition. For \texttt{InstructBLIP-vicuna-7b} (Figure~\ref{fig:instructblip_case_study}, Steps 0--1100), CLIP scores drop from 30 to 20 with modest perplexity increases, showing a similar degradation pattern in generation capability. Although the outputs at this stage have already degenerated and contain unintelligible specific tokens (e.g., ``archivi'', ``sierp 2013''), the model still retains most of its original comprehension ability. 

\noindent{\bf Stage 2: complete collapse.} At the critical threshold (Step 5 for LLaVA, Step 1200 for InstructBLIP), both models experience sharp transitions: perplexity increases dramatically while CLIP scores either become NaN or drop to around 19, meaning that the models fail to produce meaningful output (e.g., ``and, and ...'', ``ÄÄÄÄÄÄ...''). \bluetext{Due to the continuous feedback loops inherent in autoregressive generation, once the number of masked neurons exceeds the critical threshold, the accumulation of errors within the residual stream exceeds the network's ability to self-correct. The numerical shift destabilizes 
the model's internal representations and damages the generation 
process.} Continued masking beyond this threshold produces minimal 
additional change: both models stabilize in collapsed state 
and cannot recover their functionality once collapsed.%
\footnote{Notably, multiple ablation trials of the same number of 
randomly selected non-critical neurons in both models consistently 
produce negligible impact on model performance, confirming the 
specificity of these critical neurons.}

The observation that both models show similar collapse patterns but the number of critical neurons varies across architectures aligns with prior robustness studies: LLaVA achieves 89\% attack success rates under jailbreak attacks, while InstructBLIP is more resilient at 33\%~\cite{li2025attack}.

\begin{figure}[t]
    \centering
    \begin{subfigure}[t]{0.8\linewidth}
        \centering
        \includegraphics[width=\linewidth]{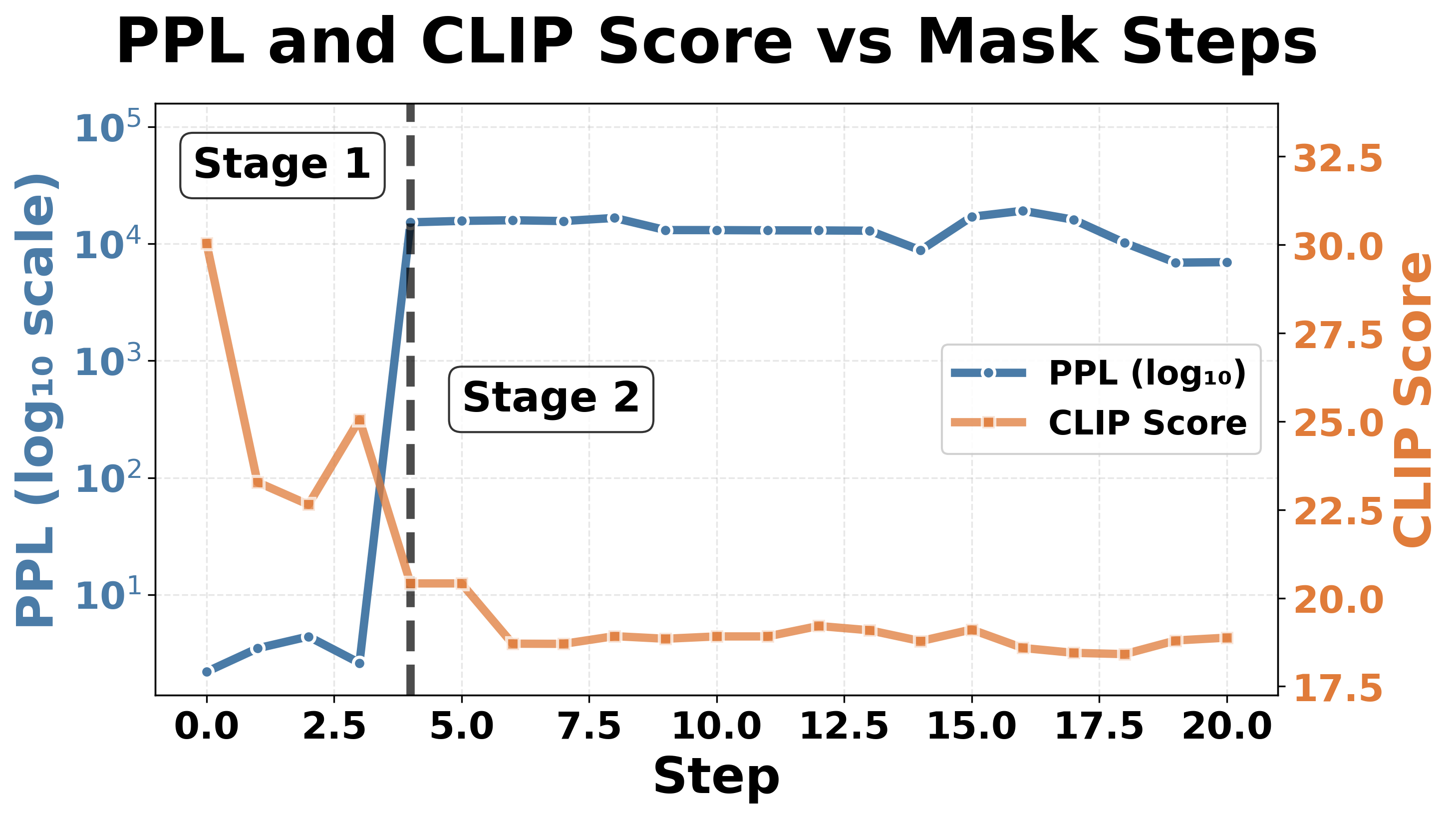}
        \caption{\texttt{LLaVA-1.5-7b-hf}}
        \label{fig:llava_case_study_downproj}
    \end{subfigure}
    
    \vspace{0.5em}
    
    \begin{subfigure}[t]{0.8\linewidth}
        \centering
        \includegraphics[width=\linewidth]{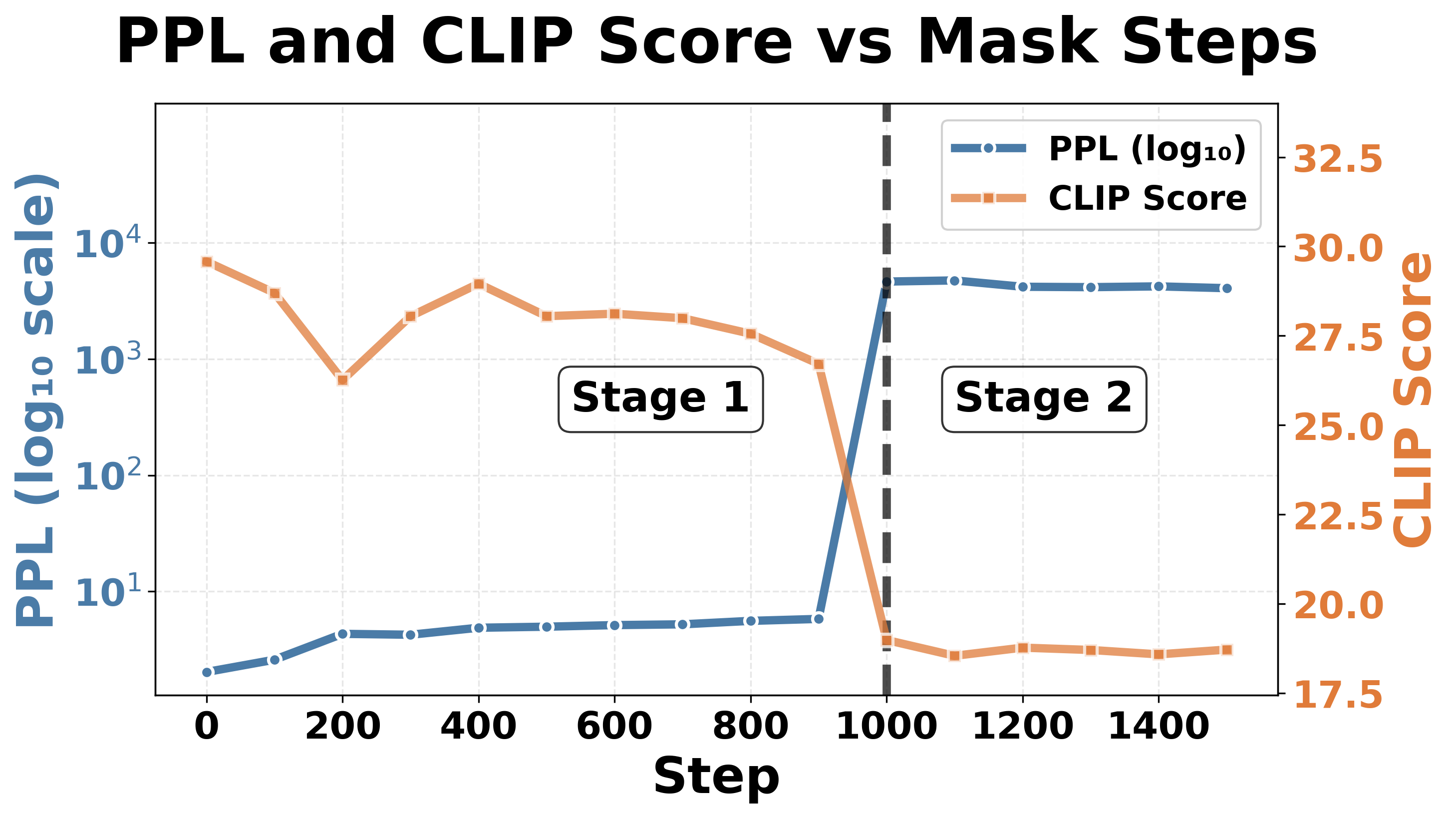}
        \caption{\texttt{InstructBLIP-vicuna-7b}}
        \label{fig:instructblip_case_study_downproj}
    \end{subfigure}
    \caption{Comparison of PPL and CLIP score changes during masking of neurons in \texttt{down\_proj} using car images from FFN for the two models. Stage 1 indicates an expressive degradation and Stage 2 indicates a complete collapse.}
    \label{fig:comparison_case_study_downproj}
\end{figure}

\subsection{Impact of mask position}
\label{sec:q3}

To assess how neuron position influences the number of critical neurons required to trigger catastrophic collapse, we conducted a comparative analysis across two distinct positions in the FFN architecture. In Section~\ref{sec:Q1}, we  examined \texttt{gate\_proj} (blue component in Figure~\ref{fig:mlp}, 11,008 dimensions), which operates on the expanded intermediate representation before activation function, and \texttt{down\_proj} (orange component in Figure~\ref{fig:mlp}, 4,096 dimensions), which performs dimensionality reduction at the network output. 
In general, masking neurons in \texttt{down\_proj} requires fewer 
critical neurons to induce collapse compared to \texttt{gate\_proj} 
(see Table~\ref{tab:collapse_downproj}). For \texttt{LLaVA-1.5-7b-hf}, 
only four neurons in \texttt{down\_proj} (compared to five in 
\texttt{gate\_proj}) are sufficient to trigger catastrophic failure, 
with perplexity increases reaching up to four orders of magnitude. 
These four neurons also remain consistent regardless of object category, 
with two neurons located in layer 1 and two neurons in layer 30 (out of 
32 layers, zero-indexed), further confirming the importance of neurons 
at the language model's boundaries. This sensitivity can be attributed 
to the dimensionality reduction performed by \texttt{down\_proj}, 
which projects the expanded intermediate representation back into the 
original dimensionality of the residual stream. \bluetext{We hypothesize 
that the identified critical neurons are indispensable within this 
projection: when masked, the rank structure of the projection space is 
disrupted, preventing high-dimensional features from mapping back into 
the residual space. Consequently, this local perturbation at a critical 
information bottleneck propagates directly into the primary information 
flow,} yielding higher post-collapse perplexity values than those observed 
under \texttt{gate\_proj} masking (Table~\ref{tab:collapse_comparison}). 
Similarly, \texttt{InstructBLIP-vicuna-7b} shows the same pattern of 
increased vulnerability in \texttt{down\_proj}.

Masking critical neurons in \texttt{down\_proj} for both models results in severe output corruption. For \texttt{LLaVA-1.5-7b-hf}, masking \texttt{down\_proj} produces empty outputs with only end-of-sentence tokens, while masking \texttt{gate\_proj} generates repetitive character strings, as shown in Figure~\ref{fig:corrupted_output_example}. In contrast, \texttt{Instruct-BLIP-vicuna-7b} outputs repetitive tokens after masking both positions, suggesting model-specific vulnerability patterns to neuron masking in FFN components. Figure~\ref{fig:comparison_case_study_downproj} confirms the same 
two-stage collapse pattern for \texttt{down\_proj}, showing a stronger bottleneck than 
\texttt{gate\_proj} due to its role in information synthesis.%
\footnote{Repeated ablations of the same number of randomly 
selected non-critical neurons in both \texttt{down\_proj} 
components consistently produce negligible impact, further 
confirming the specificity of these critical neurons.}

\begin{figure}[t]
    \centering
    \begin{subfigure}[t]{0.75\linewidth}
        \centering
        \includegraphics[width=\linewidth]{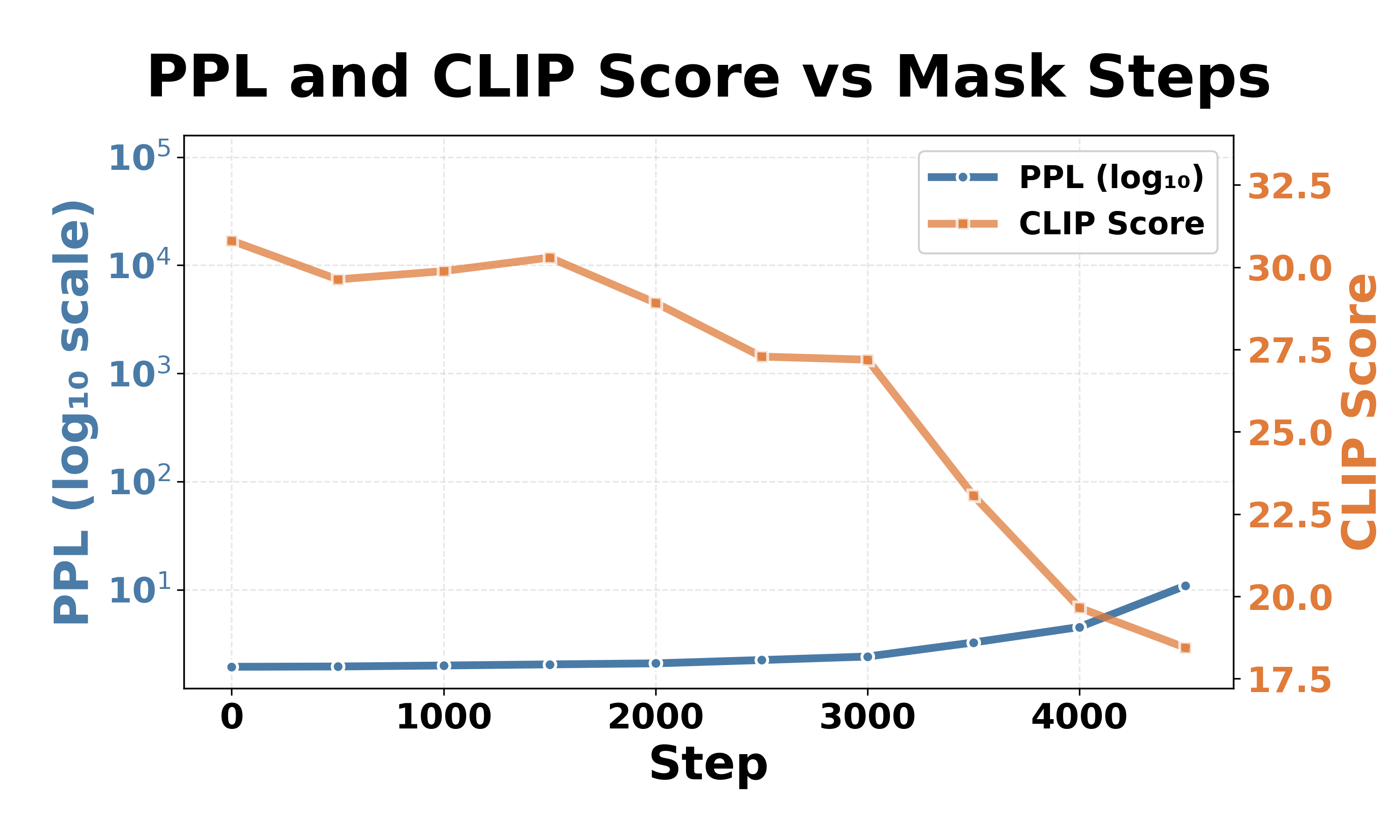}
        \caption{\texttt{LLaVA-1.5-7b-hf}}
        \label{fig:llava_mmproj}
    \end{subfigure}
    
    \vspace{0.3em}
    
    \begin{subfigure}[t]{0.75\linewidth}
        \centering
        \includegraphics[width=\linewidth]{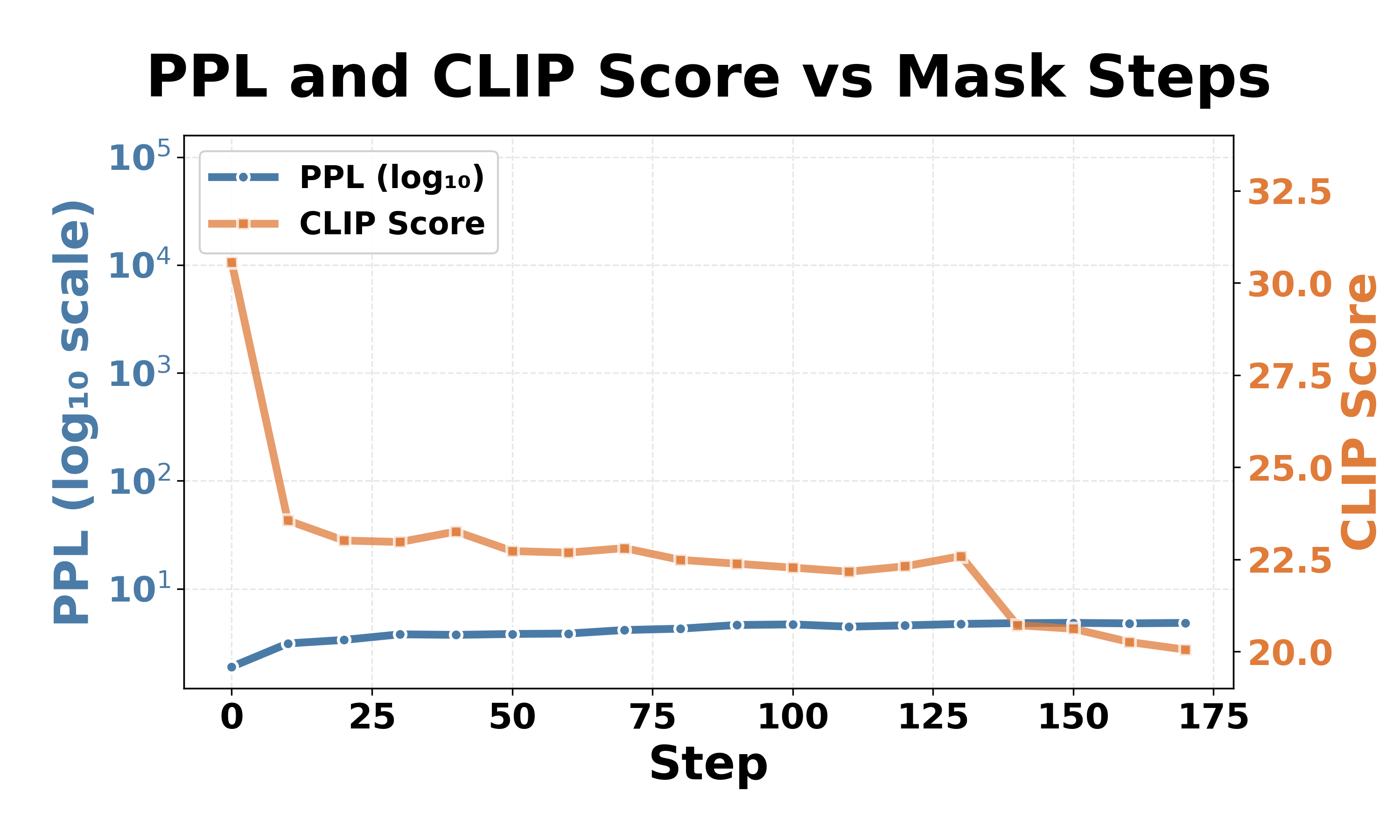}
        \caption{\texttt{InstructBLIP-vicuna-7b}}
        \label{fig:instructblip_qformer}
    \end{subfigure}
    \caption{Comparison of PPL and CLIP score changes using car images during progressive masking of neurons from \texttt{multimodal\_projector} in \texttt{LLaVA-1.5-7b-hf} and from \texttt{Q-Former} in \texttt{InstructBLIP-vicuna-7b}. In both cases, CLIP Score drops while PPL remains relatively stable, without order-of-magnitude increase.}
    \label{fig:comparison_qformer_mmprojector}
\end{figure}

\subsection{Critical neuron location}

\label{sec:critical neuron location}

\begin{table*}[h]
\centering
\caption{Ablation results for masking critical neurons in different components of \texttt{LLaVA-1.5-7b-hf}. Results are averaged over 100 test images per object category. We use steps of 500 neurons to identify approximate number of neurons in vision-related neurons leading to \emph{perceptual failure}, while PPL remains relatively stable without order-of-magnitude increase. }
\label{tab:component_ablation_llava}

\setlength{\tabcolsep}{3pt}
\renewcommand{\arraystretch}{1.0}

{\fontsize{14pt}{18pt}\selectfont
\begin{adjustbox}{width=1\textwidth}
\begin{tabular}{ccccccc}
\toprule
\multirow{2}{*}{Object} & \multicolumn{3}{c}{Multimodal Projector (4096)$^{\dagger}$} & \multicolumn{3}{c}{Vision Encoder (98304)$^{\dagger}$} \\
\cmidrule(lr){2-4} \cmidrule(lr){5-7}
& \#Neurons (\%)$^{\ddagger}$ & PPL Change ($\downarrow$) & CLIP Score Change ($\uparrow$) & \#Neurons (\%)$^{\ddagger}$ & PPL Change ($\downarrow$) & CLIP Score Change ($\uparrow$)\\
& & (Orig. $\rightarrow$ Masked$^{\S}$, \textbf{Change}) & (Orig. $\rightarrow$ Masked$^{\S}$, \textbf{\% Change}) & & (Orig. $\rightarrow$ Masked$^{\S}$, \textbf{Change}) & (Orig. $\rightarrow$ Masked$^{\S}$, \textbf{\% Change}) \\
\midrule
dog & 3500 (85\%) & 2.20 $\rightarrow$ 5.96 (\textbf{$\times$3}) & 30.35 $\rightarrow$ 20.95 (\textbf{-31\%}) & 17500 (17\%) & 2.20 $\rightarrow$ 7.30 (\textbf{$\times$3}) & 30.35 $\rightarrow$ 21.91 (\textbf{-28\%}) \\
cat & 3500 (85\%) & 2.54 $\rightarrow$ 9.83 (\textbf{$\times$4}) & 31.44 $\rightarrow$ 21.03 (\textbf{-33\%}) & 18500 (18\%) & 2.54 $\rightarrow$ 8.93 (\textbf{$\times$4}) & 31.44 $\rightarrow$ 20.17 (\textbf{-36\%}) \\
car & 4000 (97\%) & 2.26 $\rightarrow$ 4.52 (\textbf{$\times$2}) & 30.80 $\rightarrow$ 19.65 (\textbf{-36\%}) & 18500 (18\%) & 2.26 $\rightarrow$ 4.14 (\textbf{$\times$2}) & 30.80 $\rightarrow$ 19.21 (\textbf{-38\%}) \\
pencil box & 3500 (85\%) & 3.25 $\rightarrow$ 4.27 (\textbf{$\times$1}) & 29.52 $\rightarrow$ 21.16 (\textbf{-28\%}) & 30000 (30\%) & 3.25 $\rightarrow$ 4.46 (\textbf{$\times$1}) & 29.52 $\rightarrow$ 21.12 (\textbf{-28\%}) \\
sofa & 4000 (97\%) & 2.13 $\rightarrow$ 8.56 (\textbf{$\times$4}) & 31.18 $\rightarrow$ 20.17 (\textbf{-35\%}) & 30000 (30\%) & 2.13 $\rightarrow$ 8.13 (\textbf{$\times$4}) & 31.18 $\rightarrow$ 19.92 (\textbf{-36\%}) \\
table & 4000 (97\%) & 2.08 $\rightarrow$ 3.34 (\textbf{$\times$2}) & 29.72 $\rightarrow$ 21.38 (\textbf{-28\%}) & 19500 (19\%) & 2.08 $\rightarrow$ 3.42 (\textbf{$\times$2}) & 29.72 $\rightarrow$ 21.68 (\textbf{-27\%}) \\
train & 2500 (61\%) & 2.80 $\rightarrow$ 7.60 (\textbf{$\times$3}) & 28.16 $\rightarrow$ 21.11 (\textbf{-25\%}) & 19000 (19\%) & 2.80 $\rightarrow$ 4.78 (\textbf{$\times$2}) & 28.16 $\rightarrow$ 20.03 (\textbf{-29\%}) \\
apple & 4000 (97\%) & 2.08 $\rightarrow$ 9.38 (\textbf{$\times$5}) & 31.64 $\rightarrow$ 21.29 (\textbf{-33\%}) & 30000 (30\%) & 2.08 $\rightarrow$ 8.23 (\textbf{$\times$4}) & 31.64 $\rightarrow$ 21.40 (\textbf{-32\%}) \\
orange & 4000 (97\%) & 2.38 $\rightarrow$ 4.93 (\textbf{$\times$2}) & 32.06 $\rightarrow$ 20.62 (\textbf{-36\%}) & 19000 (19\%) & 2.38 $\rightarrow$ 4.60 (\textbf{$\times$2}) & 32.06 $\rightarrow$ 21.48 (\textbf{-33\%}) \\
cup & 4000 (97\%) & 2.99 $\rightarrow$ 6.70 (\textbf{$\times$2}) & 30.46 $\rightarrow$ 21.06 (\textbf{-31\%}) & 30000 (30\%) & 2.99 $\rightarrow$ 5.41 (\textbf{$\times$2}) & 30.46 $\rightarrow$ 21.98 (\textbf{-28\%}) \\
\bottomrule
\multicolumn{7}{l}{$^{\dagger}$Total neuron candidates in each component.} \\
\multicolumn{7}{l}{$^{\ddagger}$Percentages are calculated relative to the total candidates.} \\
\multicolumn{7}{l}{$^{\S}$Orig. and Masked denote metrics before and after masking critical neurons.}
\end{tabular}
\end{adjustbox}
}
\end{table*}

\begin{table*}[t]
\centering
\caption{Ablation results for masking critical neurons in different components of \texttt{InstructBLIP-vicuna-7b}. Results are averaged over 100 test images per object category. We have used steps of 500 neurons for vision encoder and steps of 10 neurons in Q-Former to identify the number of neurons leading to \emph{perceptual failure}, while PPL remains stable without order-of-magnitude increase.}

\label{tab:component_ablation_instructblip}

\setlength{\tabcolsep}{3pt}
\renewcommand{\arraystretch}{1.0}

{\fontsize{14pt}{18pt}\selectfont
\begin{adjustbox}{width=1\textwidth}
\begin{tabular}{ccccccc}
\toprule
\multirow{2}{*}{Object} & \multicolumn{3}{c}{Q-Former (9216)$^{\dagger}$} & \multicolumn{3}{c}{Vision Encoder (239616)$^{\dagger}$} \\
\cmidrule(lr){2-4} \cmidrule(lr){5-7}
& \#Neurons (\%)$^{\ddagger}$ & PPL Change ($\downarrow$) & CLIP Score Change ($\uparrow$) & \#Neurons (\%)$^{\ddagger}$ & PPL Change ($\downarrow$) & CLIP Score Change ($\uparrow$)\\
& & (Orig. $\rightarrow$ Masked$^{\S}$, \textbf{Change}) & (Orig. $\rightarrow$ Masked$^{\S}$, \textbf{\% Change}) & & (Orig. $\rightarrow$ Masked$^{\S}$, \textbf{Change}) & (Orig. $\rightarrow$ Masked$^{\S}$, \textbf{\% Change}) \\
\midrule
dog & 70 (0.76\%) & 1.75 $\rightarrow$ 4.40 (\textbf{$\times$3}) & 31.44 $\rightarrow$ 21.90 (\textbf{-30\%}) & 20000 (8.35\%) & 1.75 $\rightarrow$ 3.55 (\textbf{$\times$2}) & 31.44 $\rightarrow$ 21.79 (\textbf{-31\%}) \\
cat & 150 (1.63\%) & 1.68 $\rightarrow$ 4.09 (\textbf{$\times$2}) & 29.95 $\rightarrow$ 21.81 (\textbf{-27\%}) & 35000 (14.61\%) & 1.68 $\rightarrow$ 3.32 (\textbf{$\times$2}) & 29.95 $\rightarrow$ 21.38 (\textbf{-29\%}) \\
car & 140 (1.52\%) & 2.02 $\rightarrow$ 4.79 (\textbf{$\times$2}) & 29.58 $\rightarrow$ 20.71 (\textbf{-30\%}) & 26500 (11.06\%) & 2.02 $\rightarrow$ 3.42 (\textbf{$\times$2}) & 29.58 $\rightarrow$ 21.77 (\textbf{-26\%}) \\
pencil box & 30 (0.33\%) & 1.74 $\rightarrow$ 4.82 (\textbf{$\times$3}) & 29.68 $\rightarrow$ 22.15 (\textbf{-25\%}) & 29500 (12.31\%) & 1.74 $\rightarrow$ 3.62 (\textbf{$\times$2}) & 29.68 $\rightarrow$ 21.89 (\textbf{-26\%}) \\
sofa & 110 (1.19\%) & 1.75 $\rightarrow$ 4.24 (\textbf{$\times$2}) & 32.67 $\rightarrow$ 21.66 (\textbf{-34\%}) & 31000 (12.94\%) & 1.75 $\rightarrow$ 3.41 (\textbf{$\times$2}) & 32.67 $\rightarrow$ 21.86 (\textbf{-33\%}) \\
table & 90 (0.98\%) & 2.06 $\rightarrow$ 4.66 (\textbf{$\times$2}) & 30.99 $\rightarrow$ 21.97 (\textbf{-29\%}) & 49000 (20.45\%) & 2.06 $\rightarrow$ 5.00 (\textbf{$\times$2}) & 30.99 $\rightarrow$ 21.93 (\textbf{-29\%}) \\
train & 80 (0.87\%) & 2.34 $\rightarrow$ 6.04 (\textbf{$\times$3}) & 29.74 $\rightarrow$ 21.32 (\textbf{-28\%}) & 32500 (13.56\%) & 2.34 $\rightarrow$ 3.67 (\textbf{$\times$2}) & 29.74 $\rightarrow$ 21.78 (\textbf{-27\%}) \\
apple & 110 (1.19\%) & 1.66 $\rightarrow$ 3.78 (\textbf{$\times$2}) & 32.30 $\rightarrow$ 21.11 (\textbf{-35\%}) & 62500 (26.09\%) & 1.66 $\rightarrow$ 3.51 (\textbf{$\times$2}) & 32.30 $\rightarrow$ 21.75 (\textbf{-33\%}) \\
orange & 70 (0.76\%) & 1.65 $\rightarrow$ 4.49 (\textbf{$\times$3}) & 31.49 $\rightarrow$ 20.95 (\textbf{-33\%}) & 38000 (15.86\%) & 1.65 $\rightarrow$ 3.85 (\textbf{$\times$2}) & 31.49 $\rightarrow$ 21.10 (\textbf{-33\%}) \\
cup & 50 (0.54\%) & 1.64 $\rightarrow$ 4.07 (\textbf{$\times$2}) & 32.21 $\rightarrow$ 20.39 (\textbf{-37\%}) & 41500 (17.32\%) & 1.64 $\rightarrow$ 3.25 (\textbf{$\times$2}) & 32.21 $\rightarrow$ 21.89 (\textbf{-32\%}) \\
\bottomrule
\multicolumn{7}{l}{$^{\dagger}$Total neuron candidates in each component.} \\
\multicolumn{7}{l}{$^{\ddagger}$Percentages are calculated relative to the total candidates.} \\
\multicolumn{7}{l}{$^{\S}$Orig. and Masked denote metrics before and after masking critical neurons.}
\end{tabular}
\end{adjustbox}
}
\end{table*}

We now mask neurons in vision-related components: \texttt{multimodal\_projector} and vision encoder in \texttt{LLaVA-1.5-7b-hf}, and Q-Former and vision encoder 
in \texttt{InstructBLIP-vicuna-7b} (Tables~\ref{tab:component_ablation_llava} and~\ref{tab:component_ablation_instructblip}), revealing a distinct failure mode from language model components (Tables~\ref{tab:collapse_comparison} and 
\ref{tab:collapse_downproj}): masking vision components drops CLIP scores while perplexity remains stable, whereas masking language FFN neurons causes \emph{complete collapse}. This dissociation suggests that vision component failures result in \emph{perceptual failure}: the model loses visual grounding while retaining its linguistic ability. According to previous research on compositional reasoning and object hallucination, CLIP scores in the 15-22 range represent a ``chance-level'' baseline where outputs become semantically indistinguishable from random image-text pairings~\cite{clipscore,radford2021learningtransferablevisualmodels,yuksekgonul2022and}. This threshold is supported by our language model results, where all masked outputs converge to this range despite generating non-meaningful text. We therefore adopt 22 as the CLIP score threshold for \emph{perceptual failure}. For \texttt{LLaVA-1.5-7b-hf}, \emph{perceptual failure} requires masking 61--97\% of the multimodal projector and 17--30\% of the vision encoder, far exceeding the sparse critical neuron rates in language model FFN components. As Figure~\ref{fig:llava_mmproj} shows, even when CLIP scores drop below 22, perplexity remains stable. The vision encoder of \texttt{InstructBLIP-vicuna-7b} shows similar patterns, requiring 8--26\% masking. However, the Q-Former behaves distinctly: only 0.33--1.63\% of neurons trigger \emph{perceptual failure}, yet perplexity changes remain minimal (Figure~\ref{fig:instructblip_qformer}). This suggests that vision-related components function as perceptual pathways rather than core reasoning components: the masking causes visual impairment while preserving the model's fundamental capabilities.
In contrast, masking critical neurons in the language model's FFN components causes \emph{complete collapse} (Tables \ref{tab:collapse_comparison} and \ref{tab:collapse_downproj}): simultaneous perplexity explosion and severe CLIP score degradation indicate catastrophic failure in both comprehension and generation.

\section{Conclusion}

We investigated the vulnerabilities of LVLMs through neuron-level ablation studies. We proposed CAN, a neuron identification method combining activation magnitude with gradient sensitivity, and showed that masking as few as four neurons suffices to collapse a 7B LVLM. We identified a consistent two-stage collapse pattern and localized the vulnerability bottleneck to the language model's feed-forward networks rather than vision components. This finding challenges the intuition that failures in multimodal tasks should stem from vision-language alignment components~\cite{mindthegap}, revealing instead that the language backbone constitutes the primary structural vulnerability in LVLMs. From a mechanistic interpretability standpoint, our study exposes latent structural vulnerabilities in LVLMs, contributing empirical grounding for safety-aware interpretability research in multimodal models. Future work will extend our analysis to larger models and diverse architectures, and design defense strategies such as redundant neuron pathways.

\section{Limitation}

While our findings reveal that LVLM functionality depends on a sparse set of neurons concentrated in the language backbone, we acknowledge several limitations 
that present opportunities for future research. First, our 
investigation is primarily empirical. Our CAN method identifies 
a sufficient set of neurons that can trigger the collapse, 
but it does not guarantee that the discovered set is minimal and could be data-dependent. 
Second, conducting mechanistic analysis on LVLMs presents greater 
challenges compared to unimodal LLMs. The lack of architectural 
standardization across multimodal models, such as diverse vision 
encoders and varying cross-modal projectors, renders code 
implementation complex and model-dependent. Finally, due to 
resource constraints, our experiments were restricted to the 7B 
parameter scale. We welcome future studies to scale these ablation 
studies to larger models to verify whether massive parameter 
scaling (e.g. 70B) impacts the fragility of the language 
core in LVLMs.

\bibliography{example_paper}
\bibliographystyle{icml2026}





\end{document}